# Weighted Conformal LiDAR-Mapping for Structured SLAM

Natalia Prieto-Fernández, Sergio Fernández-Blanco, Álvaro Fernández-Blanco,
José Alberto Benítez-Andrades, Francisco Carro-De-Lorenzo, and Carmen Benavides

*Abstract*— One of the main challenges in simultaneous localization and mapping (SLAM) is real-time processing. High-computational loads linked to data acquisition and processing complicate this task. This article presents an efficient feature extraction approach for mapping structured environments. The proposed methodology, weighted conformal LiDAR-mapping (WCLM), is based on the extraction of polygonal profiles and propagation of uncertainties from raw measurement data. This is achieved using conformal M bius transformation. The algorithm has been validated experimentally using 2-D data obtained from a low-cost Light Detection and Ranging (LiDAR) range finder. The results obtained suggest that computational efficiency is significantly improved with reference to other state-of-the-art SLAM approaches.

## I. INTRODUCTION

**A**N AUTONOMOUS robotic unit must be capable of moving in a potentially unknown environment. Consequently, it requires a navigation subsystem based on three distinct steps: mapping, location, and navigation [1]. This process can be summarized in the concept simultaneous localization and mapping (SLAM). SLAM combines data from available sensors, maps the environment, and locates the robotic unit within it. The process is completed by the estimation of uncertainties, both of the mapped common features and of the unit itself.

Manuscript received 7 April 2023; revised 15 May 2023; accepted 24 May 2023. Date of publication 8 June 2023; date of current version 21 June 2023. This work has been supported by TRESCA Ingeniería S.A. within the framework of the Research Project on Survey Systems for Hostile Structured Environments using Simultaneous Localization And Mapping (SLAM) Systems with Lidar Technology. The project has been co-financed by the European Regional Development Fund (ERDF), Thematic Objective 1, which seeks to promote technological development, innovation and quality research. The Associate Editor coordinating the review process was Dr. Yan Zhuang. *(Corresponding author: Natalia Prieto-Fernández.)*

Natalia Prieto-Fernández is with the SECOMUCI Research Group, Department of Electric, Systems and Automatics Engineering, University of León, Campus of Vegaza s/n, 24071 León, Spain (e-mail: nprif@unileon.es).

Sergio Fernández-Blanco is with the Department of Electric, Systems and Automatics Engineering, University of León, Campus of Vegaza s/n, 24071 León, Spain (e-mail: sfernb03@estudiantes.unileon.es).

Álvaro Fernández-Blanco is with the Complutense University of Madrid, 28040 Madrid, Spain (e-mail: alvarf15@ucm.es).

José Alberto Benítez-Andrades and Carmen Benavides are with SALBIS Research Group, Department of Electric, Systems and Automatics Engineering, University of León, Campus of Vegazana s/n, 24071 León, Spain (e-mail: jbena@unileon.es; carmen.benavides@unileon.es).

Francisco Carro-De-Lorenzo is with TRESCA Ingeniería S.A., 24009 León, Spain (e-mail: fcarro@tresca.es).

Digital Object Identifier 10.1109/TIM.2023.3284143

In this context, cumulative errors act as a stark constraint to the quality of the navigation step. A proof of success in this step is the capacity to return to the starting position; a problem known as loop closure detection [2], [3], [4].

Light Detection and Ranging (LiDAR) sensors are currently gaining momentum in SLAM applications. They are range finders capable of determining distances based on the time of flight of a laser pulse [5]. The increased volume and robustness of data, along with its improved accuracy, are the main reasons for choosing LiDAR sensors [5], [6]. In any case, these sensors are still insufficient to become the only source of data for SLAM. Therefore, it is usual to resort to sensor fusion.

We can name inertial measurement units (IMUs) or global navigation satellite systems (GNSSs) as examples of auxiliary technologies to complement LiDAR in SLAM applications [3].

We identify two key aspects in the implementation of LiDAR SLAM algorithms: uncertainty propagation and the processing of high volumes of data. These issues, specially the latter, become increasingly critical in real-time applications with computational and energy constraints. Regarding uncertainty obtention, the algorithm must transform LiDAR 2D data into a collection of uncertainty matrices associated with the obtained natural landmarks [7]. Deriving these matrices from raw sensor data may occasionally be problematic. However, the viability of the system depends entirely on the reliability of the generated map. Offsets in position are directly linked to the precision of the calculated references. The mapping step in SLAM algorithms requires the processing of large volumes of data. Ideally, larger numbers of measurements would result in better performance. Unfortunately, this is antagonistic to resource constraints expectable in real robotic units. Thus, computational efficiency is of paramount importance. The current tendency is to extract map features directly from raw data [7]. The nature of the surroundings plays an important role in the complexity of the mapping process. We may distinguish structured, generally indoor environments in which the dominant natural features are straight lines, corners, and curves from unstructured environments with higher irregularity [7], [8]. Extrapolating, structured SLAM places lower computational strain on the algorithm than unstructured SLAM.

In this work, we propose a new formulation, based on conformal transformation, that allows for the extraction of the features characterizing the profile of a structured environment. The profile is eventually defined as a combination of straight lines and representative points, with their associated uncertainties. A LiDAR sensor provides the raw measurements required for profile characterization and uncertainty propagation. The main contribution of this work lies in the reduction of the computational cost associated with the mapping step.

This article is organized as follows. Section II provides a review of related work. In Section III, the mathematical introduction of the new model is presented. Section IV shows the details of our methodology for feature extraction. LiDAR experiments are presented in Section V. Finally, Section VI concludes this article.

## II. RELATED WORK

High data volumes obtained from LiDAR sensors allow for high-precision mapping. However, this comes at the expense of large computational efforts [6]. The current state of the art is to condense this raw data into representative features of the mapped environment. Memory consumption and data-handling benefit significantly from this approach [9], particularly in structured SLAM.

Straight line characterization can be achieved using two parameters. Arras and Siegwart [10] extract the slope and normal distance of the line containing the segment under study, starting from data obtained in polar coordinates and considering angle magnitudes so as to avoid ambiguity. Vandorpe et al. [11] characterize the line using its slope and intersection and perform lineal regression on both axes. Thus, they avoid numeric overflow caused by tangents in the vicinity of $\pi/2$. Taylor and Probert [12] and Siadat et al. [13] conceptualize the line as a linear expression defined by variable coefficients and propose global least squares fitting. Several authors, such as Nuñez et al. [7], Jensfelt and Christensen [14] or Fortin et al. [15] use the line-fitting procedure described in [10] without weighting. Sack and Burgard [16] add the expectation about each measurement belongs to line base on expectation-maximization (EM) algorithm as weighting factor. Yan et al. [17] use a factor equal to the inverted sum of the angle and distance variances. This last approach is not useful in the case of LiDAR sensors, where angle and distance uncertainties are specified by the manufacturer as constants.

Corners are also relevant geometric characteristics of a structured environment. They are of particular importance in the positioning step. The existing literature suggests several approaches to tackle corner identification. The most common procedure, used by Nuñez et al. [7], Yan et al. [17] and Castellanos et al. [18] is to characterize corners as the intersection of two consecutive straight lines. The simplest approach is to identify a corner every time there is a sign change in the distance difference between two consecutive points [19]. However, this may lead to spurious detections. Some authors, such as Amri et al. [20], suggest using the cosine theorem as a solution.

The methodology proposed here is based on the extraction of polygonal profiles, whose intersection defines keypoints (corners). In this sense, it is comparable to the approaches developed in [10] and [13]. We will, therefore, use these methodologies as reference for comparisons. Uniformity requires that we complete certain aspects of the formulation described in [10] and [13] (e.g. weighting factors). This is done following procedures found in the literature or in our own developments. The algorithms are later implemented and used in real environments. We identify a significant reduction of processing times, which necessarily translates into a faster mapping phase.

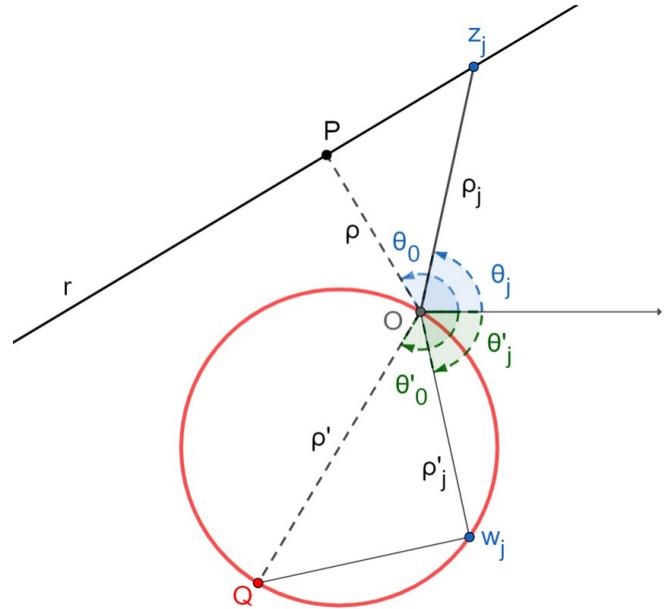

Fig. 1. Conformal transformation of line $r$ in plane $Z$ to its inverted red circumference counterpart in plane $W$.

## III. MATHEMATICAL MODEL

Angles and distances provided by LiDAR can be naturally transferred to a 2-D polar diagram. The origin of the representation is located at the instrument position; the polar axis is fixed to the instrument direction. In the complex plane, every obtained measurement can be represented by a module (distance) and a phase (angle). It is important to remark that the sensor cannot provide measurements corresponding to a line passing through the origin.

The mathematical basis of the proposed mapping approach is the inversion in the complex plane ($C$). It is an application $f : C \mapsto$ that transforms lines that do not pass through the origin into circumferences that do. It is a particular case of the Möbius transformation, to which the properties of conformity and isogonality are inherent [21]. Consequently, angles and rotation directions are kept from the complex plane $Z$ to its inverted counterpart $W$. This behavior reduces the complexity of discerning between concavity and convexity.

Fig. 1 shows point $z_j$ belonging to line $r$ in $Z$. The result of the inversion of $z_j$ in plane $W$ is point $w_j$. The locus of the inverted points of $r$ is represented in red, corresponding to a circumference passing through the origin $O$ (common to both $Z$ and $W$). The inversion of point $P$, closest point of $r$ to the origin, is $Q$. The segment $\overline{OQ}$ is seen under an angle of $\pi/2$ from all possible iterations of $w_j$. Therefore, all segments $\overline{Ow_j}$ are perpendicular to their counterparts $\overline{Qw_j}$. Please note that, for each point $z_j$, the inverse of the module is equal to the module of its inverse. In a real application, any point $z_j$ has an associated measurement uncertainty. Consequently, inverted points $w_j$ will also present dispersion. Relative uncertainties are maintained between pairs of inverted and non-inverted points. Least-squares fitting allows for the calculation of point $Q$.

Let us consider point $z_j = |z_j| e^{i\theta_j}$, with an associated inverted counterpart $w_j = (1/|z_j|) e^{-i\theta_j}$. For any value of $j$, the lines containing segments $Qw_j$ can be expressed as

$$|w_Q| \cos \left( \theta'_j - \theta'_0 \right) = |w_j| \tag{1}$$

which leads to

$$x_Q \cdot x_j - y_Q \cdot y_j = 1 \tag{2}$$

where $x^Q$ and $y^Q$ are the coordinates of point $Q$ in plane $W$ and $x_j$ and $y_j$ are the coordinates of point $z_j$ in plane $Z$.

## IV. METHODOLOGY

Let us now define the procedure to obtain poligonal characteristics and their associated uncertainties in structured environments. We define a novel methodology, weighted conformal LiDAR-mapping (WCLM), based on the inversion model defined previously. The LiDAR sensor is placed on a rotatory base. The system measures two magnitudes: distance (provided directly by the LiDAR sensor) and angle (obtained from an encoder). Since we deal with uncoupled sensors, we expect their uncertainties to be uncorrelated. Furthermore, we assume them as Gaussian with null mean. They are defined by a bivariate distribution [22] of the form

$$f(\varepsilon_\rho, \varepsilon_\theta) = \frac{1}{2\pi\sigma_\rho \rho \sigma_\theta} e^{-\frac{1}{2}\left(\frac{\varepsilon_\rho^2}{\sigma_\rho^2} + \frac{\varepsilon_\theta^2}{(\rho\sigma_\theta)^2}\right)} \tag{3}$$

where $\varepsilon_\rho$ and $\varepsilon_\theta$ are the Gaussian uncertainties associated with the distance ($\rho$) and angle ($\theta$), respectively; and $\sigma_\rho^2$ and $\sigma_\theta^2$ are

a global term simplifies the handling of singularities (i.e., vertical or horizontal lines)

$$\varepsilon_j = x_Q \cdot \varepsilon_{x_j} - y_Q \cdot \varepsilon_{y_j}. \tag{6}$$

Let us now consider a set of sample points associated with line $r$, written in matrix form

$$\begin{bmatrix} x_i & y_i & 1 \\ x_{i+1} & y_{i+1} & 1 \\ x_{i+2} & y_{i+2} & 1 \\ \cdots & \cdots & \cdots \\ x_n & y_n & -1 \end{bmatrix} \cdot \begin{bmatrix} x_Q \\ -y_Q \\ 1 \end{bmatrix} = \begin{bmatrix} \varepsilon_i \\ \varepsilon_{i+1} \\ \varepsilon_{i+2} \\ \cdots \\ \varepsilon_n \end{bmatrix}. \tag{7}$$

which can be expressed as

$$\mathbf{H} \cdot \mathbf{V} = \boldsymbol{\varepsilon} \tag{8}$$

where $\mathbf{H}$ is the observation matrix and $\mathbf{V}$ is the vector of coordinates of point $Q$. Minimizing the weighted error vector

$$V^T \cdot H^T \cdot W \cdot H \cdot V = \sum_{i=1}^{n} w_i \varepsilon_i^2 \tag{9}$$

where $W = P^T P$ is the diagonal matrix which contains the weighting terms $\omega_i$. Each of these terms can be defined as

$$\omega_i = \frac{1}{V_{x_i} V_{y_i} - C_{x_i y_i}^2}. \tag{10}$$

With $V_{x_i}$ and $V_{y_i}$ representing the conditioned variances in both axes, with covariance $C_{x_i, y_i}$. The terms correspond to

the inverse of the dispersion in the bivariate distribution in Cartesian coordinates, given by

$$f \qquad \frac{1}{e^k} \qquad (11)$$

their variances. Please note that there are no cross-correlated terms $C_{(\varepsilon_\rho, \varepsilon_\theta)}$, due to the aforementioned independence of the sensors.

$$2\pi\sigma_{\varepsilon_x}\sigma_{\varepsilon_y}$$

$$(\varepsilon_x, \varepsilon_y) = \qquad 1 - p^2$$

$$= [\qquad] \quad = [\qquad]$$
$$= [\qquad]$$

weighted by the value of parameters $x_Q$ and $y_Q$ . Usage of

Based on the assembly described above, we perform the where $k = -(1/(2(1-p^2)))[(\varepsilon^2/\sigma^2) + (\varepsilon^2/\sigma^2)$

—

$$\left\{\left(\begin{array}{l}?\\?\end{array}\right)\right\}! \cap \epsilon_{\sigma_{\varepsilon_x}\sigma_{\varepsilon_y}}$$

$$p^2 = -\frac{C^2_{\varepsilon_x \varepsilon_y}}{\sigma^2_{\varepsilon_x} \sigma^2_{\varepsilon_y}}. \tag{12}$$

Terms $V_{x_i}$, $V_{y_i}$ and $C_{x_i y_i}$, from covariance matrix $C_{p_i}$, can be obtained from

$$\begin{bmatrix} V_{x_i} & C_{x_i y_i} \end{bmatrix} = \mathbf{J_{xy}} \cdot \begin{bmatrix} V_{\rho_i} & C_{\rho_i \theta_i} \end{bmatrix} \cdot \mathbf{J^T}$$

(13)

*1) WCLM for Line Fitting:* Parameters $x_Q$ and $y_Q$ which

=

define line $r$, can be obtained using least-squares fitting. We achieve this by minimizing the

$C_{x_i y_i}$ $V_{y_i}$ $C_{\rho_i \theta_i}$ $V_{\theta_i}$ **xy**

Euclidean norm of the

where $V_{\rho_i}$ and $V_{\theta_i}$ are the distance and angle dispersions,

respectively, and $C_{i,\rho_i,\theta_i}$, coordinate change, defined by

0. $J_{xy}$ is the Jacobian of the

$$J_{xy}$$

$$= \begin{bmatrix} \end{bmatrix} \partial \rho$$

$C_{x_i y_i}$     $V_{y_i}$       $T_{21}$     $T_{22}$

$$\frac{\partial y_j}{\partial \theta_i}$$

$\partial \theta_i$

$$\begin{bmatrix} V_{x_i} & C_{x_i y_i} \end{bmatrix} = \begin{bmatrix} T_{11} & T_{12} \end{bmatrix} \tag{15}$$

where

$$T_{11} = V_{\rho_i} \cos^2 \theta_i + \rho_i^2 V_{\theta_i} \sin^2 \theta_i \tag{16a}$$

$$T_{12} = V_\rho - \rho^2 V_\theta \sin \theta_i \cos \theta_i \tag{16b}$$

$$T_{22} = V_\rho \sin^2 \theta_i + \rho_i^2 V_{\theta_i} \cos^2 \theta_i. \tag{16c}$$

By substituting terms, we reach the same weighting factor (17) found by operating in polar coordinates. Therefore, we find it to be independent of the coordinate frame

$$\omega_i = \frac{1}{\sigma^2 (\rho \sigma_\theta)^2}. \tag{17}$$

We derive the following expressions for $x_Q$ and $y_Q$:

$$x_Q$$

$$= \frac{\sum_i w_i x_i \left( \sum_i w_i y_i^2 - \sum_i w_i y_i \right) \sum_i w_i x_i y_i}{\left( \sum w_i x^2 \right) \left( \sum w_i y^2 \right) - \left( \sum w_i x_i y_i \right)^2} \tag{18}$$

$$y_Q$$

$$= \frac{\sum w_i x_i \left( \sum w_i x_i y_i \right) - \sum w_i y_i \left( \sum w_i x^2 \right)}{\left( \sum w_i x^2 \right) \left( \sum w_i y^2 \right) - \left( \sum w_i x_i y_i \right)^2}. \tag{19}$$

In which $A_i$, $B_i$, $C_i$, $D_i$, $R_i$, and $M$ can be calculated as

$$A_i = x_i \overline{Y^2} - y_i \overline{XY} \tag{25a}$$

$$B_i = y_i \overline{X^2} - x_i \overline{XY} \tag{25b}$$

$$C_i = R_i \left( y_i \overline{Y^2} + x_i \overline{XY} \right) + \rho_i^2 \overline{Y} \tag{25c}$$

$$D_i = R_i \left( x_i \overline{X^2} + y_i \overline{XY} \right) + \rho_i^2 \overline{X} \tag{25d}$$

$$R_i = 1 - \left( \frac{2 x_i x_Q - 2 y_i y_Q}{M} \right) \frac{1}{2} \tag{25e}$$

where average values are weighted using factor $\omega_i$ (25f)

*2) Arras–Siegwart Method for Line Fitting:* The parameters used in [10] to define a line are $r$, distance of the line to the origin of coordinates, and $\alpha$, angle between the normal to $r$ and the polar axis

$$r = \frac{\sum_i w_i \rho_i \cos(\theta_i - \alpha)}{\sum w_i} \tag{26}$$

polar parameters $(\rho_i, \theta_i)$ defined for $n$ points of line $r$.

$$\tan(2\alpha)$$

$$= \frac{\sum_i \sum_j w_i \rho_i w_j \rho_j \sin \theta_i \cos \theta_j - \sum_i w_i \rho^2 \sin 2\theta_i}{\sum_i \sum_j w_i \rho_i w_j \rho_j \cos(\theta_i + \theta_j) - \sum_i w_i \rho^2 \cos 2\theta_i}$$

$C_{x\ y}$    $V_y$     $_{\varrho Q}$    $C_{\rho\theta}$    $V_\theta$     $_{x_Q\,y_Q}$

$\times$

$$C_Q = \sigma_\rho \cdot \quad \underline{\quad x \quad y \quad} \quad \partial y \quad ^2 \qquad\qquad\qquad _{\text{line}} \quad C_{r\alpha} \quad V_r \quad ^{r\alpha}$$

$$+ \sigma^2 \cdot \quad \partial\theta_i \qquad \partial\theta_i \ \partial\theta_i \quad \qquad (21).$$

$$C_{\text{line}} = \sigma_\rho^2 \cdot \quad \frac{\underline{\partial a}\ ^2 \quad \underline{\partial a}\ \partial r}{r}$$

$(\sigma_\rho\ ,\ \sigma_\theta\ )$. We have obtained the estimators

Uncertainty propagation from sensor data to the line con-

where

$$C_{Q_{1i}} = \qquad M\rho_i \qquad - A_i B_i \qquad B_i^2 \qquad\qquad (23)$$

the implicit representation of the straight line in the following

for point $Q$ from

siders the uncertainty values provided by the manufacturer

$$\alpha = \frac{1}{} \tan^{-1} \quad \underline{N} \qquad (28)$$

$$(27)$$

*i*

$$C_Q = \begin{bmatrix} V_{x_Q} & C_{x_Q y_Q} \\ & \end{bmatrix} = J_{xy} \cdot \begin{bmatrix} V_\rho & \\ & C_{\rho\theta} \end{bmatrix} \cdot J^T$$

$$\tag{20}$$

The development of this

methodology in [10] leaves an

open

$$C_{\rho\theta} \quad \sigma^2 \quad ^{r a}$$

$$\frac{\boxed{i}}{F_i} \quad {}_2\square$$

$$\square \qquad \frac{\partial\theta\,\square}{\partial\theta_i}\,{}_2\square$$

choice regarding the weighting

coefficients. Some alternatives

are provided, but these are not

easily accessible parameters in

where $J_{x_Q y_Q}$ is the $2 \times 2n$ Jacobian matrix of the coordinates of point $Q\,(x_Q,\ y_Q)$ with respect to sensor data $(\rho_i,\theta_i)$. The following uncertainty matrix results:

$$\left(\frac{\partial x_Q}{\ }\right)^2 \qquad \frac{\partial x_Q\ \partial y_Q}{\ }$$

the case of low-cost instruments. Here, we will consider the inverse of the deviation associated with a bivariate distribution,

as defined in (17), in analogy to the methodology WCLM

followed in Section IV-A1. Parameters $r$ and $a$ are not

uncorrelated. Their uncertainty matrix [10] is defined by

2 $\square$

$$C \begin{bmatrix} V_a \\ \sigma_2 \end{bmatrix} \quad C_{ra} = J \begin{bmatrix} C_{\rho\theta} \end{bmatrix} \cdot J^T$$

$i$ $\square$ $\partial_Q$ $\partial_Q$ $Q$ $\square$ $\theta$

$\partial \rho_i$ $\partial \rho_i$ $\partial \rho_i$

where covariance $C_{\rho\theta}$ is zero, $J_{r\alpha}$ is the Jacobian matrix of

$\square$

$\partial x_Q$ $^2$ $\partial x_Q \, \partial y_Q$ $\square$

line     p   arameters $C(r,\ a)$          with respect to sensor data $(\rho_i,\ \theta_i)$

$\theta$      $\partial x_O\ \partial y_O$      $\underline{\partial y_O}$   $^2$

$$\square \frac{\partial a}{\partial} \quad \partial \quad \square$$

$$i \qquad i \qquad i$$

Which we shall abbreviate as

$$\frac{\partial \rho}{\partial \rho_i} \quad {}_i$$

$$_2 \quad \frac{\partial a}{}$$

$$\partial a \ \partial r$$

$\partial \rho_i$

$i$        $i$

$$+ \sigma_\theta \cdot \frac{\square}{\square} \frac{}{\partial \sigma \;\; \partial r}$$

$$-\omega_i \quad {}^2 \quad A^2 \qquad -A_i B_i$$

*3) Siadat Method for Line Fitting:* Siadat et al. [13] use

$$\frac{\omega_i}{C^2}$$

2 .

$C_i$ $D_i$

$$(24)$$

equation to

define the

unweighted

line:

$M$     $C_i D_i$     $D_i$

$$ax + by + c = 0. \tag{31}$$

The weighting factor $\omega_i$ applied in (31) is defined in (17). To extract the parameters $a$, $b$, and $c$, the Lagrange Multiplier method [22] has been used, subjected to the constraint: $a^2 + b^2 = 1$. Parameters $a$–$c$ are not uncorrelated. Their uncertainty matrix is

$$C_{abc} = \begin{bmatrix} V_a & C_{ab} & C_{ac} \\ C_{ab} & V_b & C_{bc} \\ C_{ac} & C_{bc} & V_c \end{bmatrix} = J_{abc} \cdot \begin{bmatrix} \sigma_\rho^2 & C_{\rho\theta} \\ C_{\rho\theta} & \sigma_\theta^2 \end{bmatrix} \cdot J_{abc}^T \tag{32}$$

where covariance $C_{\rho\theta}$ is zero and $J_{abc}$ is the Jacobian matrix of line parameters $C(a, b, c)$ with respect to sensor data $(\rho_i, \theta_i)$.

$$C_{abc} = \sigma_\rho^2 \cdot \begin{bmatrix} \left(\frac{\partial a}{\partial \rho_i}\right)^2 & \frac{\partial a}{\partial \rho_i}\frac{\partial b}{\partial \rho_i} & \frac{\partial a}{\partial \rho_i}\frac{\partial c}{\partial \rho_i} \\ \frac{\partial a}{\partial \rho_i}\frac{\partial b}{\partial \rho_i} & \left(\frac{\partial b}{\partial \rho_i}\right)^2 & \frac{\partial b}{\partial \rho_i}\frac{\partial c}{\partial \rho_i} \\ \frac{\partial a}{\partial \rho_i}\frac{\partial c}{\partial \rho_i} & \frac{\partial b}{\partial \rho_i}\frac{\partial c}{\partial \rho_i} & \left(\frac{\partial c}{\partial \rho_i}\right)^2 \end{bmatrix}$$

$$+ \sigma_\theta^2 \cdot \begin{bmatrix} \left(\frac{\partial a}{\partial \theta_i}\right)^2 & \frac{\partial a}{\partial \theta_i}\frac{\partial b}{\partial \theta_i} & \frac{\partial a}{\partial \theta_i}\frac{\partial c}{\partial \theta_i} \\ \frac{\partial a}{\partial \theta_i}\frac{\partial b}{\partial \theta_i} & \left(\frac{\partial b}{\partial \theta_i}\right)^2 & \frac{\partial b}{\partial \theta_i}\frac{\partial c}{\partial \theta_i} \\ \frac{\partial a}{\partial \theta_i}\frac{\partial c}{\partial \theta_i} & \frac{\partial b}{\partial \theta_i}\frac{\partial c}{\partial \theta_i} & \left(\frac{\partial c}{\partial \theta_i}\right)^2 \end{bmatrix} \tag{33}$$

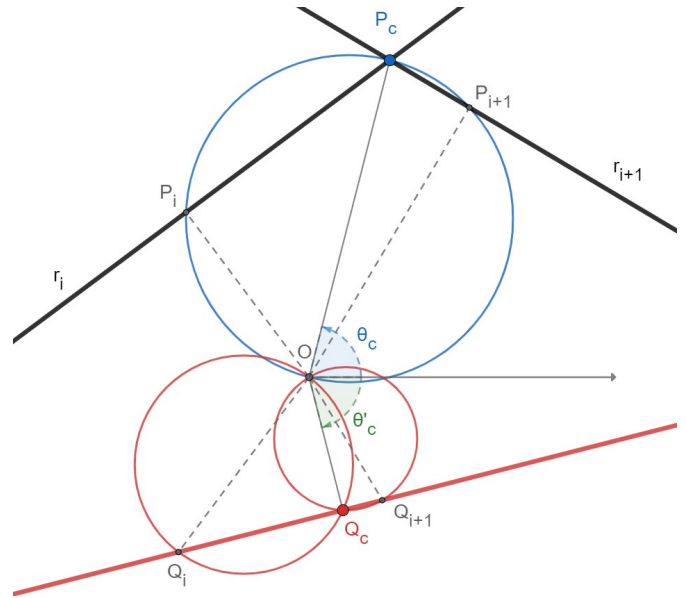

Fig. 2. Extraction of corners as the intersection of the inverse red circumferences of the straight lines $r_i$ and $r_{i+1}$, in plane W. The intersection point obtained is $Q_c$, the inverse of $P_c$.

where

$$J_{P_c} = \begin{bmatrix} \frac{\partial x_{Q_i}}{\partial x_{P_c}} & \frac{\partial y_{Q_i}}{\partial y} & \frac{\partial x_{Q_{i+1}}}{\partial y} & \frac{\partial y_{Q_{i+1}}}{\partial x_{P_c}} \end{bmatrix}.$$

### B. Corner Characterization

$\square$  $P_c$

$P_c$

$P_c$

Solving the system, we calculate the coordinates of the

$$C_c = \begin{matrix} x_c & x_c y_c \\ C_{x_c y_c} & V_{y_c} \end{matrix} = J_c \cdot C_{r_i q_{i n_{1} i}} q_i \cdot J_c^T \qquad (42)$$

$$\overline{\qquad x \qquad}$$

given by

$$\begin{matrix} c & c & & & & P_c^T \\ & & 0 & 0 & C_{xy_{Q_{i+1}}} & V_{y_{Q_{i+1}}} \end{matrix}$$

*1) WCLM for Corner Extraction:* The coordinates of point $Q$ are sufficient to define line $r$. We can define two lines $r_i$ and

$r_{i+1}$, characterized by points $Q_i$ and $Q_{i+1}$. The circumferences

in $W$ resulting from the inversion of $r_i$ and $r_{i+1}$ intersect at point $Q_c$. This point is the inverse of the intersection point of lines $r_i$ and $r_{i+1}$ in $Z\,(P_c)$ and is contained by segment

$Q_i\,Q_{i+1}$. In fact, we   can invert the blue circumference  found        in Fig. 2, which verifies

$$\partial x_{Q_{i+1}}$$

$$
_c \boxminus
\begin{array}{cccc}
0 & 0 & V_{r_{i+1}} & C_{r_{i+1}a_{i+1}} \\
0 & 0 & C_{r_{i+1}a_{i+1}} & V_{a_{i+1}} \\
& & & \partial a_{i+1} \\
\partial r_i & & & \partial a_{i+1}
\end{array}
$$

*2) Corner Extraction Based on Arras–Siegwart:* The parameters defined in [10] serve as the basis for the corner

lines $r_i$ and $r_{i+1}$

characterization procedure proposed in [7]. Considering two

$$r_i = x_c \cos(a_i) + y_c \sin(a_i)$$

$$r_{i+1} = x_c \cos(a_{i+1}) + y_c \sin(a_{i+1}). \tag{39}$$

The coordinates of $P_c$ are found by solving the system

$$x \quad \frac{r_i \sin(a_{i+1}}{\sin(a_{i+1} - a_i)}$$

$$y \quad \frac{r_{i+1}\cos(a_i) - r_i\cos(a_{i+1})}{\sin(a_{i+1} - a_i)}. \qquad (41)$$

The uncertainty propagation of line parameters to corner

corner $(x_{P_c}, y_{P_c})$

parameters is reflected in matrix $C_c$

$$V \quad C$$

$$r_i \boldsymbol{a}_i r_{i+1} \boldsymbol{a}_{i+1}$$

$$x_{P_c} = x_Q \cdot y_{Q_{i+1}} - y_{Q_i} \tag{36}$$

parameters of the two lines and $J_c$ is the Jacobian matrix of

$$\partial a_i \qquad \partial r_{i+1}$$

corner parameters with respect to line parameters

$$\cdot x_Q$$

$$y_{P_c} = \qquad - x_{Q_i} \qquad - y_Q \qquad \cdot x_Q$$

$$x_Q \cdot y_Q^{Q_{i+1}}$$

$$\square \quad V_{r_i} \qquad \mathcal{G}_i \qquad \begin{pmatrix} \\ \end{pmatrix}$$

$$. \qquad (37)$$

$$C = \begin{bmatrix} C_{r\,a} & V_a & 0 & 0 \end{bmatrix}$$

$r_i a_i r_{i+1} a_{i+1}$

The covariance matrix that determines corner uncertainty is

$$\begin{bmatrix} V_{x_{Q_i}} & C_{xy_{Q_i}} & 0 & 0 \end{bmatrix} \qquad \begin{bmatrix} \dfrac{\partial x_c}{} & \dfrac{\partial x_c}{} & \dfrac{\partial x_c}{} & \dfrac{\partial x_c}{} \end{bmatrix}$$

$xy_{Q_i}$    $V_{y_{Q_i}}$    $J$    $\partial r_i$    $\partial r_{i+1}$    $\square$.    (43b)

$C_P = J_P$    $C$    $0$    $0$    (38)    $\partial a_i$

$\cdot \square$    $0$    $0$    $V_{x_{Q_{i+1}}}$    $C_{xy_{Q_{i+1}}} \square \cdot J$    $\square \underline{\partial v_i}$

$$\frac{}{\partial \gamma_c} \; \square$$

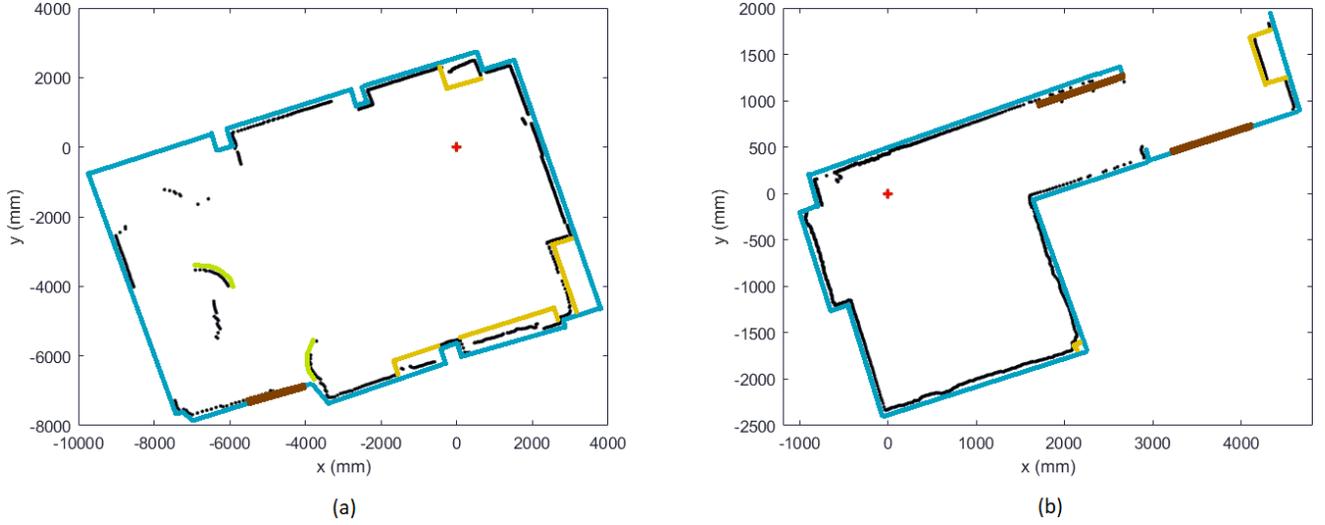

Fig. 3. Raw data supplied by the RPLiDAR S1 sensor. The sensor position is represented by the red cross while black dots show the mapped environment data. The real environment is furnished with cabinets, some of which are glazed, represented in yellow, brown lines show doors, while the green areas correspond to curved cabinets. (a) Environment A. (b) Environment B.

*3) Corner Extraction Based on Siadat:* The corner extraction is based on the intersection of two straight lines obtained by Siadat

$$a_i x_c + b_i y_c + c_i = 0$$
$$a_{i+1} x_c + b_{i+1} y_c + c_{i+1} = 0. \quad (44)$$

Solving the system, the corner coordinates $(x_c, y_c)$ are obtained

$$x_c = \frac{b_i c_{i+1} - b_{i+1} c_i}{a_i b_{i+1} - a_{i+1} b_i} \quad (45)$$

$$y_c = \frac{a_{i+1} c_i - a_i c_{i+1}}{a_i b_{i+1} - a_{i+1} b_i}. \quad (46)$$

Matrix $C_c$ reflects the uncertainty propagation of the line parameters to the corner parameters

$$C_c = \begin{matrix} V_{x_c} & C_{x_c y_c} \\ C_{x_c y_c} & V_{y_c} \end{matrix} = J_c \cdot C_{a_i b_i c_i a_{i+1} b_{i+1} c_{i+1}} \cdot J_c^T \quad (47)$$

where $C_{a_i b_i c_i a_{i+1} b_{i+1} c_{i+1}}$ is the covariance matrix associated with the parameters of the two lines and $J_c$ is the Jacobian matrix of corner parameters with respect to line parameters

$$C_{a_i b_i c_i a_{i+1} b_{i+1} c_{i+1}} = \begin{matrix} C_{a_i b_i c_i} & 0 \\ 0 & C_{a_{i+1} b_{i+1} c_{i+1}} \end{matrix} \quad (48)$$

where

$$J_c = \begin{bmatrix} \frac{\partial x_c}{\partial a_i} & \frac{\partial x_c}{\partial b_i} & \frac{\partial x_c}{\partial c_i} & \frac{\partial x_c}{\partial a_{i+1}} & \frac{\partial x_c}{\partial b_{i+1}} & \frac{\partial x_c}{\partial c_{i+1}} \\ \frac{\partial y_c}{\partial a_i} & \frac{\partial y_c}{\partial b_i} & \frac{\partial y_c}{\partial c_i} & \frac{\partial y_c}{\partial a_{i+1}} & \frac{\partial y_c}{\partial b_{i+1}} & \frac{\partial y_c}{\partial c_{i+1}} \end{bmatrix}$$

## V. EXPERIMENTAL RESULTS

The device employed for the acquisition of the indoor datasets is RPLIDAR S1 by SLAMTEC. It is an omnidirectional laser range scanner. Its angular resolution is ($\sigma_\theta = 0.391/2°$), and its depth accuracy, ($\sigma_\rho = \pm 5$ cm). These values condition the uncertainty of the obtained profiles. The measurements are represented in polar coordinates on a flat section of the environment, centered around the sensor.

TABLE I
CORNER UNCERTAINTIES OF ENVIRONMENT A IN mm: WCLM,
ARRAS–SIEGWART, AND SIADAT

| Corner Number | WCLM | | Arras-Siegwart | | Siadat | |
|---|---|---|---|---|---|---|
| | $\sigma_{x,P}$ | $\sigma_{y,P}$ | $\sigma_{x,P}$ | $\sigma_{y,P}$ | $\sigma_{x,P}$ | $\sigma_{y,P}$ |
| 1 | 71.72 | 69.58 | 61.55 | 69.98 | 87.63 | 56.91 |
| 2 | 29.88 | 35.64 | 28.96 | 37.89 | 36.03 | 49.88 |
| 3 | 21.60 | 25.06 | 21.45 | 25.54 | 32.16 | 45.40 |
| 4 | 392.09 | 117.96 | 385.77 | 117.00 | 274.85 | 118.57 |
| 5 | 32.73 | 32.07 | 32.40 | 32.37 | 40.54 | 42.49 |
| 6 | 73.05 | 162.50 | 75.67 | 192.46 | 93.68 | 209.37 |
| 7 | 156.04 | 158.55 | 132.87 | 141.82 | 160.02 | 197.79 |
| 8 | 65.98 | 105.50 | 62.44 | 105.96 | 90.45 | 132.62 |
| 9 | 59.37 | 160.94 | 48.04 | 147.48 | 44.93 | 169.85 |
| 10 | 79.75 | 25.53 | 87.16 | 27.39 | 92.52 | 32.75 |
| 11 | 42.60 | 67.02 | 42.80 | 66.68 | 65.58 | 68.37 |
| 12 | 24.90 | 93.16 | 27.16 | 93.29 | 31.01 | 115.01 |
| 13 | 62.28 | 49.75 | 62.14 | 49.19 | 100.78 | 67.93 |
| 14 | 153.93 | 36.35 | 147.17 | 31.85 | 199.50 | 44.78 |
| 15 | 24.71 | 64.24 | 24.64 | 63.81 | 44.38 | 46.09 |
| 16 | 21.18 | 110.98 | 21.36 | 110.12 | 39.27 | 170.16 |
| 17 | 20.25 | 130.82 | 20.60 | 129.38 | 37.87 | 202.63 |
| 18 | 25.90 | 13.21 | 26.02 | 12.45 | 46.06 | 30.00 |
| 19 | 16.48 | 15.80 | 17.08 | 16.24 | 34.07 | 39.21 |
| 20 | 13.68 | 9.58 | 13.21 | 9.16 | 32.09 | 41.62 |
| 21 | 13.79 | 14.48 | 13.26 | 14.01 | 29.06 | 39.94 |
| 22 | 10.48 | 12.12 | 9.86 | 11.48 | 39.25 | 34.35 |
| 23 | 14.43 | 23.05 | 13.60 | 22.42 | 32.98 | 45.12 |
| 24 | 13.53 | 17.31 | 14.22 | 17.69 | 25.68 | 44.66 |
| 25 | 19.71 | 76.18 | 25.36 | 97.78 | 23.72 | 92.39 |
| 26 | 28.90 | 55.67 | 24.30 | 56.57 | 49.24 | 74.57 |

Raw sensor data is grouped into straight segments. They correspond to the walls of the surveyed environment. In a first approximation, this is achieved using a classic line tracking method [13]. To associate a new point to a segment, its distance to the line is calculated. If the value lies below a threshold (20 mm), association takes place.

As shown in Fig. 3, two different scenarios have been surveyed to test the effectiveness of the proposed method (WCLM). Environment A is approximately 12 × 8 m while environment B is approximately 5×3 m. Both environments correspond to structured SLAM, with several straight segments and two curves. A LiDAR sweep in environment A yields 884 points, which conform into 54 straight segments with 26 corners. Note that curves are split into short straight segments. This is the case for the segments comprised between corners 6–9 and 13–14, see Fig. 4(a). For environment B, a LiDAR sweep delivers 1112 points, which fit into 20 straight segments with eight corners [see Fig. 4(b)]. The datasets for each indoor environment were obtained from a fixed position.

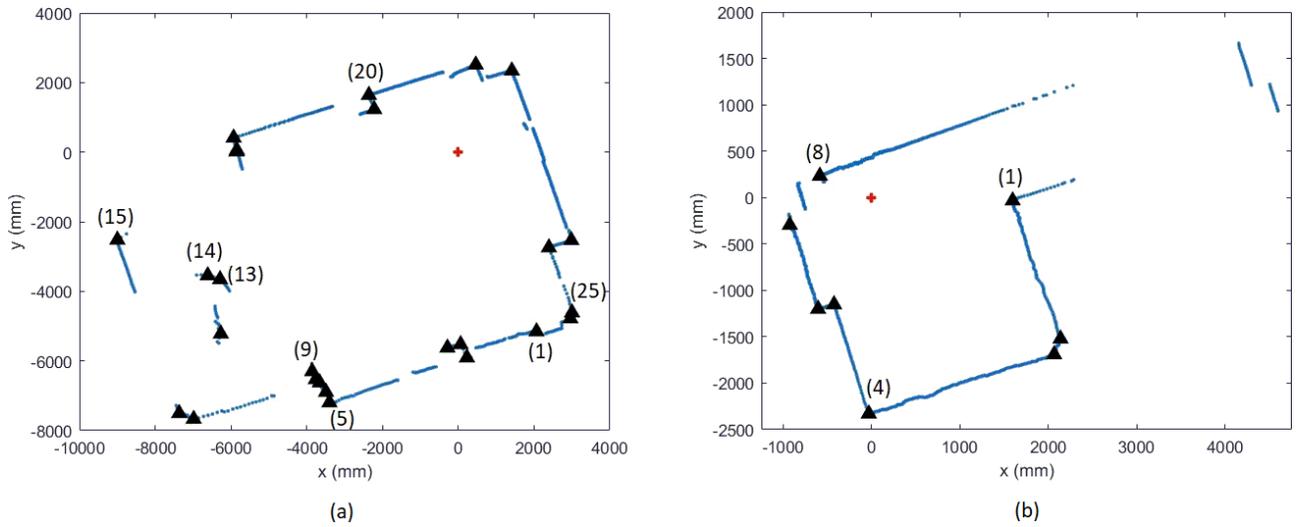

Fig. 4. Feature extraction. Corners are represented by black triangles, the sensor position by a red cross and lines obtained from the raw data are highlighted in blue. Corners of (a) environment A and (b) environment B.

TABLE II
CORNER UNCERTAINTIES OF ENVIRONMENT B IN mm: WCLM, ARRAS–SIEGWART, AND SIADAT

| Corner Number | WCLM | | Arras–Siegwart | | Siadat | |
|---|---|---|---|---|---|---|
| | $\sigma_{x_{P_c}}$ | $\sigma_{y_{P_c}}$ | $\sigma_{x_{P_c}}$ | $\sigma_{y_{P_c}}$ | $\sigma_{x_{P_c}}$ | $\sigma_{y_{P_c}}$ |
| 1 | 9.46 | 6.15 | 10.14 | 6.64 | 45.28 | 22.02 |
| 2 | 24.34 | 67.62 | 26.35 | 73.00 | 29.60 | 78.60 |
| 3 | 66.78 | 22.06 | 68.17 | 22.17 | 63.65 | 53.45 |
| 4 | 8.75 | 14.60 | 7.81 | 14.48 | 21.78 | 46.07 |
| 5 | 8.67 | 19.13 | 9.45 | 19.70 | 32.25 | 39.64 |
| 6 | 8.95 | 13.81 | 8.34 | 13.50 | 36.64 | 36.86 |
| 7 | 65.00 | 170.91 | 64.19 | 167.81 | 53.95 | 206.08 |
| 8 | 11.10 | 10.06 | 10.73 | 9.17 | 33.09 | 30.38 |

TABLE III
COMPARATIVE COMPUTATIONAL TIMES OF WCLM, ARRAS-SIEGWART, AND SIADAT IN $\mu$s: CORNER 20 OF THE ENVIRONMENT A

| Number of Points | WCLM | Arras - Siegwart | Siadat |
|---|---|---|---|
| 10 | 36.48 | 36.26 | 53.72 |
| 20 | 43.45 | 69.31 | 84.45 |
| 30 | 47.86 | 113.87 | 117.89 |
| 40 | 56.77 | 172.39 | 149.29 |
| 50 | 61.34 | 251.44 | 183.15 |
| 60 | 71.93 | 350.27 | 214.98 |
| 70 | 76.60 | 448.57 | 241.24 |
| 80 | 84.64 | 583.00 | 269.17 |
| 90 | 91.90 | 740.02 | 303.25 |
| 100 | 104.19 | 961.22 | 354.75 |
| 110 | 107.03 | 1149.36 | 397.81 |
| 120 | 119.68 | 1402.59 | 422.28 |
| 130 | 130.38 | 1611.93 | 475.65 |

TABLE IV
COMPARATIVE COMPUTATIONAL TIMES OF WCLM, ARRAS-SIEGWART, AND SIADAT IN $\mu$s: CORNER 4 OF THE ENVIRONMENT B

| Number of Points | WCLM | Arras - Siegwart | Siadat |
|---|---|---|---|
| 10 | 31.68 | 35.21 | 49.50 |
| 20 | 38.65 | 81.18 | 82.92 |
| 30 | 48.68 | 137.35 | 112.89 |
| 40 | 53.44 | 203.61 | 140.75 |
| 50 | 61.35 | 289.82 | 174.63 |
| 60 | 67.66 | 399.05 | 211.43 |
| 70 | 75.88 | 517.87 | 245.05 |
| 80 | 83.10 | 664.02 | 279.10 |
| 90 | 90.01 | 796.67 | 294.43 |
| 100 | 100.75 | 1009.51 | 343.16 |
| 110 | 104.82 | 1154.16 | 361.87 |
| 120 | 111.53 | 1352.22 | 388.26 |
| 130 | 122.04 | 1589.32 | 424.76 |
| 140 | 128.48 | 1778.76 | 476.31 |
| 150 | 133.15 | 1991.65 | 494.49 |
| 160 | 145.41 | 2261.51 | 529.53 |
| 170 | 148.12 | 2502.88 | 576.84 |
| 180 | 154.62 | 2808.24 | 599.72 |
| 190 | 168.83 | 3235.20 | 659.64 |
| 200 | 169.88 | 3549.85 | 683.86 |
| 210 | 184.84 | 3899.99 | 735.39 |
| 220 | 190.57 | 4288.27 | 775.83 |
| 230 | 196.87 | 4757.09 | 816.83 |

Tables I and II contain the processing results using the methods described above: WCLM, Arras–Siegwart, and Siadat.

Corner coordinates are very similar for all methods. The average corner uncertainties on the $x$-axis for Arras–Siegwart and WCLM are 5.6 versus 6.8 cm for Siadat. On the other hand, the average corner uncertainties of the $y$-axis are 6.5 cm for Arras–Siegwart and WCLM versus 8.5 cm for Siadat. Siadat is thus less accurate than Arras–Siegwart and WCLM, which yield similar uncertainties. The order of magnitude of these values is directly comparable to the sensor distance uncertainty (5 cm).

All algorithms have been implemented in MATLAB R2021a, running on an Intel Core i9-9900K processor CPU and 64 GB RAM. We have selected four corners (environment A: 20 and 22; environment B: 1 and 4) for performance assessment. These corners have been deemed representative of the ensemble, since all of them feature

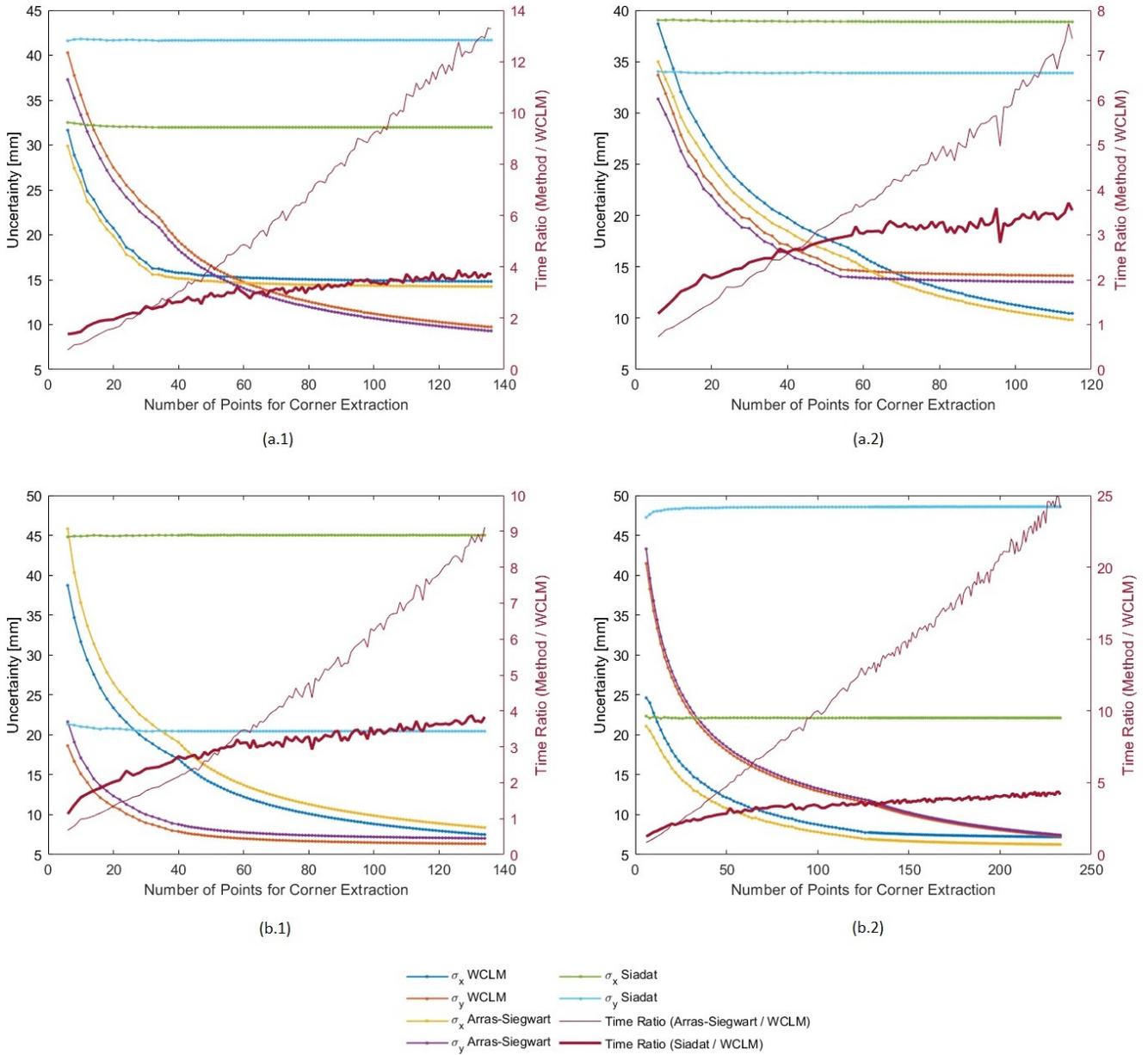

(a.1)  (a.2)

(b.1)  (b.2)

Fig. 5.  Analysis of corner parameters as a function of the number of points used in their estimation. The left axis shows the uncertainty in Cartesian coordinates of the corner for Arras–Siegwart, Siadat, and WCLM. The right axis shows the computational cost of the Arras–Siegwart and Siadat with respect to the WCLM. (a.1) Corner 20 of the environment A. (a.2) Corner 22 of the environment A. (b.1) Corner 1 of the environment B. (b.2) Corner 4 of the environment B.

equivalent behavior patterns in terms of uncertainty and computing effort. For each of them, uncertainties and required computational times (of the full algorithm) have been obtained as a function of the number of available points. The values are available for WCLM, Arras–Siegwart, and Siadat.

Fig. 5 represents the evolution of uncertainties and computational time ratio with the number of computed points. Table III contains detailed computational time results for corner 20 (environment A) while Table IV compiles the computational time for corner 4 (environment B).

The proposed method (WCLM) requires less computation time than Arras–Siegwart and Siadat. On the one hand, the Jacobian matrices used in the Arras–Siegwart for the propagation of the uncertainties from the sensor to the line parameters are computationally more complex than their WCLM counterparts. Indeed, these Jacobian matrices are obtained from double summations and inverse tangents, which results in a higher computational cost of the Arras–Siegwart algorithm. On the other hand, the Siadat requires the calculation of three parameters ($a$–$c$), their uncertainties and their propagation to the corners. These requirements are softened in WCLM, where only two parameters ($x_Q$ and $y_Q$) are used.

Regarding uncertainty levels, Fig. 5 shows Siadat to be sistematically less accurate with respect to the other analyzed methods. In the case of Arras–Siegwart and WCLM, even if uncertainties remain comparable, WCLM remains more efficient computationally as the number of points increases. Uncertainties follow an asymptotical pattern, converging at a certain value. In the "transient" phase before this stabilization happens, the computational time ratio reaches values up to 10–15 between Arras–Siegwart and WCLM, to the advantage of the latter.

As a final remark, it is worth noting that the variability in uncertainty convergence between the six combinations of figures and axes is linked to the number of available points and their position relative to the sensor. Despite an expectable dependence on the specific geometry under analysis, WCLM processing times are shown to be significantly shorter.

## VI. CONCLUSION

A new 2-D LiDAR feature extraction methodology has been presented in this article. The key contribution of WCLM to the state-of-the-art SLAM is the improved efficiency in terms of computational time, while preserving uncertainty levels. The fact that processing times show restrained growth with the number of processed points implies that higher data volumes may be managed simultaneously. Thus, the potential of certain sensors can be fully or partially unlocked, alleviating the bottleneck that the SLAM algorithm constitutes. Beside this, the fact that larger sets of measurements can be processed results in incremental uncertainty reductions in a given computation timespan.

WCLM characterizes the line by its inversion point. The associated mathematical treatment is based on global least-squares fitting and prevents typical numerical overflow when the line slope tends to infinity. The linear nature of the algorithm enables its implementation on edge devices such as FPGA or ASIC.

We propose WCLM as a competitive alternative for real-time SLAM applications. The experimental results obtained are comparable in terms of quality and robustness to those of other algorithms, incorporating non-trivial enhance- ments. Further investigations are likely to yield additional efficiency leaps, such as those associated with improved segmentation methodologies. The final objective would be to implement WCLM in the complete SLAM process, aiming to provide more accurate and efficient solutions for the deter- mination of the position and trajectory followed by a robotic element.

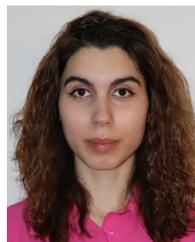

**Natalia Prieto-Fernández** received the B.S. degree in industrial, electronic and automation and the M.S. degree in industrial engineering from the University of León, León, Spain, in 2018 and 2020, respectively, where she is currently pursuing the Ph.D. degree with the SECOMUCI Research Group, Department of Electric, Systems and Automatics Engineering.

From 2019 to 2020, she worked as a Pre-Doctoral Researcher with the Research and Development Department, TRESCA Ingeniería S.A., León. Since 2020, she has been Teaching Assistant with the University of León. Her research interests include 2-D LiDAR SLAM, mobile robot navigation, and data fusion.

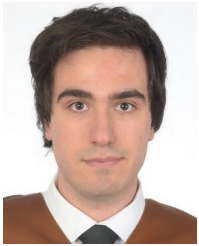

**Sergio Fernández-Blanco** received the B.S. degree in mechanical engineering and aerospace engineering from the University of León, Spain, and the M.S. degree in aeronautical engineering from the Carlos III University of Madrid, Spain, in 2018, 2019 and 2021, respectively. He is currently completing his studies in Industrial Engineering at the University of León and Space Engineering at the Carlos III University of Madrid, both at M.S. degree level. He also pursues his Ph.D. degree with the Department of Electric, Systems and Automatics Engineering of the University of León.

In 2021, he joined the Department of Flight Dynamics and Operations at GMV Aerospace and Defence. Since 2021, he works in the Flight Physics Domain of Airbus Defence and Space, in the field of static loads. His research involves uncertainty analysis within orbit propagation in the field of SSA.

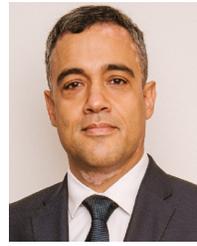

**Francisco Carro-De-Lorenzo** received the B.S. degree in industrial engineer from Comillas Pontifical University (ICAI), Madrid, Spain, in 1998, and the Ph.D. degree from the University of León, León, Spain, both in electrical power systems, in 2017.

From 2018 to 2022, he has been an Adjunct Professor of electricity generation with the University of León. He is currently the CEO of TRESCA Ingenieria S.A., León. He has more than 20 years of experience in consulting engineering and development of industrial plants. He participates as a member of various boards of directors of companies in the industrial field. He has collaborated as an expert in research projects in the area of energy and industrial plants. His research interests include hydrogen, energy storage, and digital twins in industry.

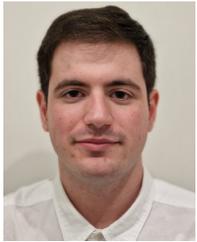

**Álvaro Fernández-Blanco** is currently pursuing the double B.S. degree in physics and mathematics with the Complutense University of Madrid, Madrid, Spain.

He was recognized as one of the highest-standing students of his generation accessing his University. His research interests include quantum computing, in which he has taken specific formation imparted by Menéndez Pelayo International University.

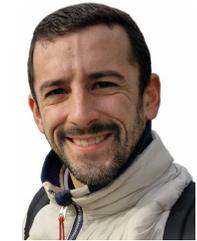

**José Alberto Benítez-Andrades** received the B.S. degree in computer engineering and the Ph.D. degree in production engineering and computing from the University of León, León, Spain, in 2010 and 2017, respectively.

He is an Associate Professor with the University of León. He has more than 40 publications indexed in JCR, 30 communications in international conferences. His research interests include the application of artificial intelligence techniques, knowledge engineering, and social network analysis applied mainly to problems related to the field of health.

Dr. Benítez-Andrades has been an Associate Editor of the journal BMC Medical Informatics and Decision Making and PeerJ Computer Science, has organized several international conferences, since 2018, and is an evaluator of international projects for the government of Spain and Peru.

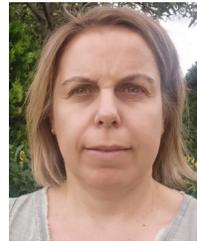

**Carmen Benavides** received the bachelor's degree in industrial technical engineer from the University of León, León, Spain, in 1996, and the master's degree in electronic engineering from the University of Valladolid, Valladolid, Spain, in 1998, and the Ph.D. degree in computer science from the University of León, in 2009.

She has been working as an Assistant Professor with the University of León, since 2001. She has organized several congresses, and has presented and published different papers in Journals, Conferences and Symposia. Her research interests include applied artificial intelligence and knowledge engineering technique to different domains.